\documentclass{article} 
\usepackage{iclr2027_conference,times}

\usepackage{amsmath,amsfonts,bm}

\def\eqref#1{equation~\ref{#1}}

\def\1{\bm{1}}

\DeclareMathAlphabet{\mathsfit}{\encodingdefault}{\sfdefault}{m}{sl}
\SetMathAlphabet{\mathsfit}{bold}{\encodingdefault}{\sfdefault}{bx}{n}

\usepackage{hyperref}
\usepackage{url}

\usepackage{amssymb} 
\usepackage{graphicx}
\usepackage{booktabs}
\usepackage{threeparttable}
\usepackage{multirow}
\usepackage{float}

\title{What Should a Navigator Remember? Spatiotemporal Memory for Zero-Shot Vision-Language Navigation}

\author{
\textbf{Yuhang Peng\textsuperscript{2}, Zhikai Wang\textsuperscript{2}, Ziyi Wang\textsuperscript{2}, Jiaru Zhong\textsuperscript{2}}\\
\textbf{Xiaoji Zheng\textsuperscript{3}, Jingzhi Huang\textsuperscript{2}, Yi Wang\textsuperscript{2}, Jiangtao Gong\textsuperscript{1}}\\
\textsuperscript{1}Shanghai Jiao Tong University
\textsuperscript{2}The Hong Kong Polytechnic University\\
\textsuperscript{3}Tsinghua University\\
}

\iclrfinalcopy 
\begin{document}
\maketitle

\fancyhead{}
\lhead{Preprint}

\begin{abstract}
Zero-shot vision-and-language navigation (VLN) requires an agent to use past observations without navigation-specific training. We study what information an 8B-scale navigation system should retain and how it should be organized for subsequent decisions. Our agent combines instruction progress, hierarchically compressed observation history anchored to poses, and accumulated geometry in complementary task, temporal, and spatial memories. An 8B observer summarizes observations, and an 8B reasoner uses the memories to select waypoint intents or request additional views, a geometric controller executes the selected targets. A semantic information perspective motivates the distinction between retaining evidence, making it accessible to the reasoner, and acquiring further observations. On R2R-CE val-unseen, the complete system achieves the highest among the evaluated Qwen3-based zero-shot baselines. Removing temporal memory reduces SR and sharply increases average path length from 16.9 to 39.1\,m, removing the spatial-map and planning component reduces SR as well. Real-world experiments on a Unitree Go2 show that our zero-shot method achieves higher success rates than the post-trained baseline in both corridor and office environments, suggesting better transfer to the evaluated real-world settings. As a complementary evaluation, LAW2 uses paired instruction interventions to examine behavioral responses, instruction following ability beyond endpoint success. These results support preserving task progress, observation history, and geometry in structured memory for zero-shot navigation with smaller open models.

\end{abstract}
\section{Introduction}

Vision-and-Language Navigation (VLN) requires an embodied agent to follow a natural-language instruction through a previously unseen environment, providing a foundation for higher-level tasks such as object manipulation and human--robot interaction. Post-training methods~\citep{guo2026awarevln, wei2026ground, wang2026lightnav} fine-tune vision-language models on navigation trajectories and achieve strong success rates on R2R-CE~\citep{krantz_vlnce_2020}. Zero-shot methods~\citep{chen2025constraint, shi2025smartway, yuan2025opennav, li2026bridging} instead reuse pretrained models without navigation-specific training. Despite their different training requirements and inference resources, both approaches must integrate visual observations, instruction progress, and execution history to make reliable navigation decisions.

A central question is how navigation history should be represented for smaller open models. In our reproduced HSGM configuration~\citep{li2026bridging}, replacing GPT-5 with Qwen3-VL-8B reduces SR from 0.463 to 0.049 on R2R-CE val-unseen (Table~\ref{tab:r2r_val_unseen}). This comparison motivates examining the demands placed on the navigation backbone. We focus on memory representation: the reasoner must connect current observations with instruction progress and earlier evidence, often across many actions.

Our design preserves three complementary kinds of information: what the instruction requires and which subgoal is active, what the agent has observed and executed, and what geometry it has explored. We organize these as task memory, temporal memory with hierarchical summaries anchored to poses, and spatial memory containing an occupancy map and a committed target.

The semantic communication perspective of~\citet{lastras2026fundamental} motivates retaining evidence useful for task-relevant judgments and organizing it for a computationally bounded receiver. In navigation, these judgments include whether a subgoal is complete and where to move next. We use this perspective as a design rationale. An 8B observer constructs observational summaries, while a Qwen3-VL-8B reasoner combines them with the instruction and spatial memory. Waypoint intents connect these decisions to geometric control, and a \texttt{PREVIEW} action allows the agent to acquire additional views.

We evaluate this memory design through navigation results, component ablations, and deployment on a Unitree Go2 to experiment real world performance. We also introduce LAW2 as a complementary diagnostic benchmark: paired direction reversals, route-information removal, and paraphrases reveal how behavior responds to instruction changes with the same meaning.

Our contributions are as follows.
\begin{itemize}
    \item We propose a spatiotemporal memory representation for zero-shot VLN that combines instruction progress, pose-anchored observation history, and accumulated geometry.
    \item We evaluate the representation in a navigation system with an 8B observer and an 8B reasoner. Component ablations characterize the effects of temporal memory, spatial memory and active previewing with planning, an integration with CA-Nav and real-world deployment provide additional evidence of utility.
    \item We introduce LAW2 to complement navigation metrics with paired tests of directional responsiveness, dependence on route information, and sensitivity to paraphrasing.
\end{itemize}

\section{Related Work}
\subsection{Vision-and-Language Navigation}
Traditional autonomous driving research has primarily focused on local obstacle avoidance and trajectory generation, rather than high-level route planning~\cite{jiang2023vad,peng2025bench2freead, zhong2025cooptrack, zheng2025coirl, yu2025end}. In contrast, Vision-and-Language Navigation (VLN) is a fundamental capability of embodied AI that enables robots to navigate autonomously by integrating natural language understanding with visual perception. The ability to accurately interpret linguistic instructions and navigate to specified target locations provides an essential foundation for more complex embodied tasks, including object manipulation and human–robot interaction.

R2R~\citep{anderson2018vision} studies navigation from natural-language instructions on a discrete graph, while R2R-CE~\citep{krantz_vlnce_2020} extends the task to continuous environments with low-level actions. REVERIE~\citep{qi2020reverie} additionally requires identifying an object referred to by the instruction~\cite{cai2024bridging, huang2026lerr, huang2026have}. These settings require agents to connect language with visual observations and spatial decisions.

Spatial representations are a central part of VLN systems. ETPNav~\citep{an2024etpnav} uses online topological mapping and obstacle-aware planning, while BEVBERT~\citep{an2022bevbert} combines global topological and local metric representations. Post-training approaches~\citep{guo2026awarevln,wei2026ground,wang2026lightnav} adapt pretrained models using navigation data. Our work studies how observation history and geometry can support navigation when the observer and reasoner are used without navigation-specific training.

\subsection{Agentic Navigation and Memory}
Agentic VLN systems combine language or vision-language models with perception, planning, and execution modules~\citep{xin2026agentvln,huang2026sedualvln,huang2026aerr,li2026agenticnav}. FreeAskWorld~\citep{peng2026freeaskworld} extends evaluation to dynamic environments and proactive tasks which needs robot to search more information when lack of task-relevant information. These settings motivate retaining information across successive decisions.

Existing systems use different representations and interfaces for navigation. CA-Nav~\citep{chen2025constraint} uses a value map to score locations by their relevance to the instruction, and BSC-Nav~\citep{ruan2026brain} explores brain-inspired spatial memory. We evaluate our approach alongside zero-shot systems including SmartWay, Open-Nav, and HSGM~\citep{shi2025smartway,yuan2025opennav,li2026bridging}. Our focus is the organization of complementary evidence: task progress, compressed observation history tied to poses, and accumulated geometry. The experiments assess this representation within a complete navigation system.

\subsection{Spatial and Temporal Memory for Navigation}

Memory representations for VLN organize both observation history and spatial structure. HAMT~\citep{chen2021history} hierarchically encodes panoramic observations and their temporal relationships for multimodal action prediction. GridMM~\citep{wang2023gridmm} projects historical observations into a unified spatial grid and aggregates visual evidence according to instruction relevance. These approaches establish the importance of structured history and spatially organized evidence. Our work examines how such information can be represented explicitly for pretrained observers and reasoners used without navigation-specific training.

Recent VLM-based navigation systems further explore explicit memory representations. MapNav~\citep{zhang2025mapnav} uses annotated semantic maps in place of historical frames, while HSGM~\citep{li2026bridging} organizes geometric, semantic, and decision information to support subtask reasoning and waypoint planning. HAM-VLN~\citep{liu2026ham} maintains a grounded world graph with decision-coupled semantic and reflective records, combining a recent-history window with subgoal-conditioned retrieval. These methods demonstrate that map construction, task progress, and selective history access are already important elements of navigation memory.

Our design emphasizes the complementary roles of temporal records and accumulated geometry for an 8B-scale reasoner.

\begin{figure}[htbp]
    \centering
    \includegraphics[width=0.9\textwidth]{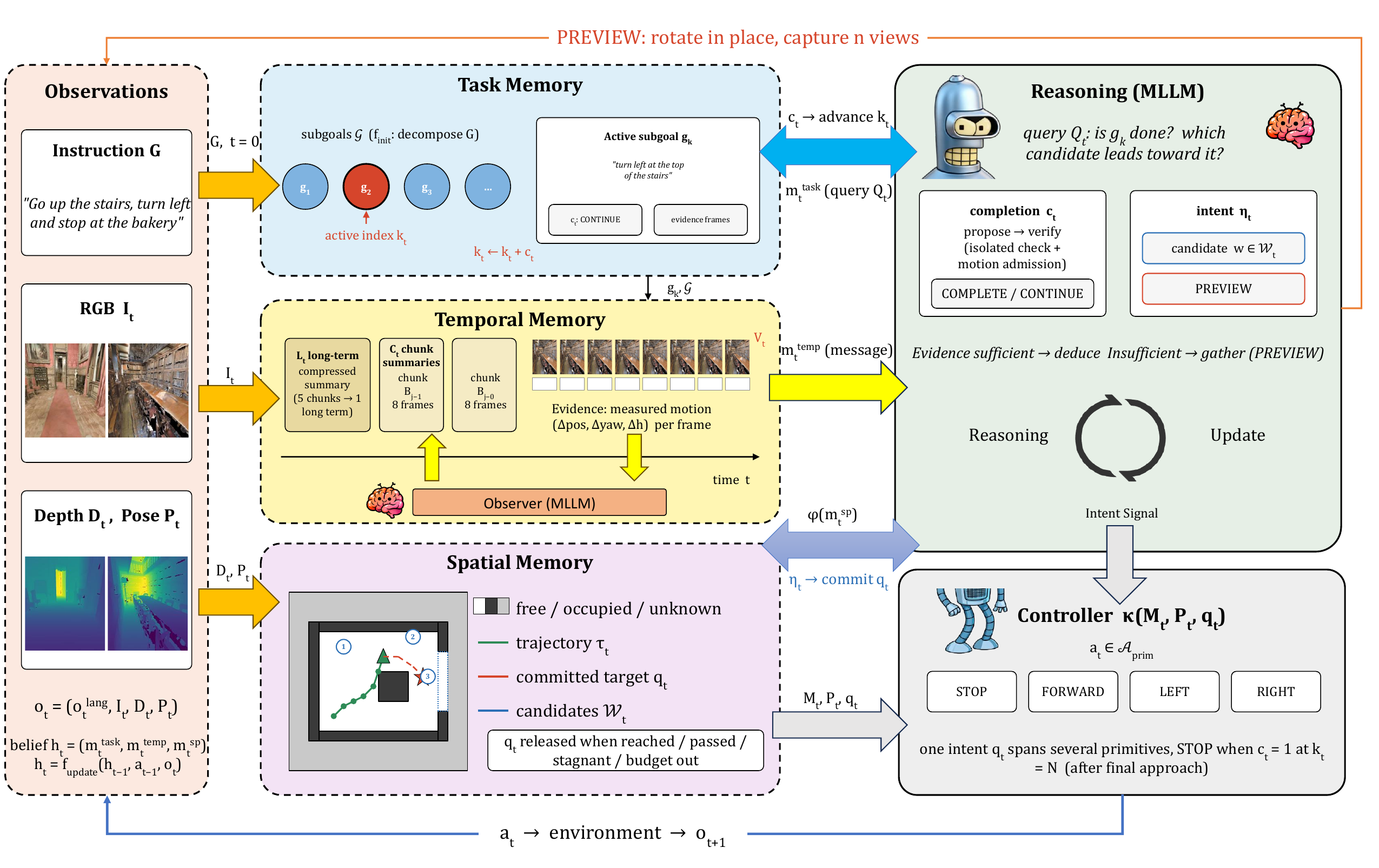}
    \caption{Spatiotemporal memory for zero-shot VLN. Task memory tracks instruction progress, temporal memory retains pose-anchored observations and summaries, and spatial memory accumulates geometry and the committed target. The reasoner uses these representations to select a waypoint or request additional views through \texttt{PREVIEW}, geometric control executes the waypoint and subsequent observations update memory.}
    \label{fig:example}
\end{figure}

\section{Methodology}
\subsection{Problem Formulation}
\label{sec:problem_formulation}

We consider vision-and-language navigation in a partially observable environment. An agent receives a natural-language instruction $G$ at the beginning of an episode and subsequently acquires observations while executing actions. We model the interaction as a POMDP~\cite{kaelbling1998planning}, whose state $s_t$ includes the environment layout and the agent's physical configuration. The agent does not observe the complete state; instead, it receives observations $v_t=(I_t,D_t,P_t)$, comprising an RGB image, depth, and a measured or estimated pose.

The environment evolves according to
\begin{equation}
s_{t+1}\sim T(\cdot\mid s_t,a_t),
\qquad
v_{t+1}\sim O(\cdot\mid s_{t+1},a_t),
\label{eq:vln_dynamics}
\end{equation}
where $T$ and $O$ are the transition and observation models. The executed action belongs to
\begin{equation}
\mathcal{A}_{\mathrm{prim}}
=
\{\texttt{FORWARD},\texttt{LEFT},\texttt{RIGHT},\texttt{STOP}\}.
\label{eq:vln_action_space}
\end{equation}

The agent maintains an internal memory that integrates the instruction with its interaction history:
\begin{equation}
\begin{aligned}
h_0 &= f_{\mathrm{init}}(G,v_0), \\
h_t &= f_{\mathrm{update}}(h_{t-1},a_{t-1},v_t),
\qquad t\geq 1, \\
a_t &\sim \pi(\cdot\mid h_t).
\end{aligned}
\label{eq:vln_memory_policy}
\end{equation}
Here, $h_t$ is an internal memory representation, not necessarily an exact Bayesian belief state, and $\pi$ denotes the complete decision-and-control policy. The instruction is supplied once, but its retained representation remains available throughout navigation.

Let $Q_G$ denote the task requirements associated with $G$. Successful execution requires the resulting trajectory to satisfy $Q_G$. Standard benchmark SR evaluates the narrower condition of stopping within a prescribed distance of the goal and does not by itself establish compliance with all route instructions. We therefore evaluate navigation performance together with instruction responsiveness.

Appendix~\ref{app:semantic_information_theory_of_embodied_ai} develops the general receiver-based perspective underlying this formulation, distinguishing available information, realized computation, and task achievement through environment interaction.

\subsection{Design Rationale}
\label{sec:design_rationale}

Our memory design is motivated by the semantic information perspective of \citet{lastras2026fundamental}, which studies communication that enables a receiver to derive required conclusions using its existing knowledge. For navigation, this motivates retaining evidence that supports task-relevant judgments rather than reproducing the entire observation stream. We adopt this perspective as a design rationale, without transferring its communication bounds directly to memory cost or navigation performance.

We distinguish evidence availability from its effective use. A navigation decision may fail because the necessary evidence has not been observed, because it was discarded during memory compression, or because the reasoner cannot reliably interpret and combine the retained information. These possibilities motivate complementary mechanisms: acquiring additional observations, retaining relevant history, and organizing that history into accessible representations. The agent is not assumed to identify these failure sources perfectly.

We instantiate this rationale through task, temporal, and spatial memory. Task memory retains the instruction and its execution stage; temporal memory organizes past observations and measured motion; spatial memory accumulates geometric evidence for navigation. Hierarchical summarization and pose anchoring organize historical information for the reasoner, while a preview mechanism allows the agent to request additional views. These mechanisms support memory-conditioned decisions without guaranteeing evidence sufficiency or correct reasoning.

Appendix~\ref{sec:semantic_information} formalizes the distinction between semantic sufficiency and realized computation. The following sections describe our memory representation and the policy that uses it to select and execute navigation actions.

\subsection{Spatiotemporal Memory}
\label{sec:spatiotemporal_memory}
We instantiate the internal memory in Eq.~(\ref{eq:vln_memory_policy}) with three complementary components:
\begin{equation}
h_t=
\left(
m_t^{\mathrm{task}},
m_t^{\mathrm{temp}},
m_t^{\mathrm{sp}}
\right),
\label{eq:structured_memory}
\end{equation}
which retain the task context, temporal observations, and accumulated spatial evidence, respectively. This representation organizes information from the interaction history without explicitly maintaining a posterior distribution over environment states. The memory update uses RGB observations $I_t$, aligned depth $D_t$, and camera poses $P_t$, with known camera intrinsics $K$.

\subsubsection{Task Memory}
Task memory stores the instruction, its ordered decomposition into
subgoals, and the active subgoal index:
\begin{equation}
    m_t^{\mathrm{task}} =
    \left(G,\mathcal{G},k_t\right),
    \qquad
    \mathcal{G}=(g_1,\ldots,g_N).
\end{equation}
The active index $k_t$ advances when the reasoning module determines
that the current subgoal has been completed. This representation
preserves route order and conditions navigation on the current
stage of the instruction.

\subsubsection{Temporal Memory}
Temporal memory represents the observed trajectory at multiple
resolutions:
\begin{equation}
    m_t^{\mathrm{temp}} =
    \left(W_t,C_t,L_t,E_t\right),
\end{equation}
where $W_t$ contains recent observations, $C_t$ stores pending
chunk-level summaries, $L_t$ contains compressed long-term summaries,
and $E_t$ records measured execution history. Recent observations are associated with frame indices, the camera pose $P_t$, and motion evidence, including translation, rotation, and elevation change.

The short-term window retains the last eight real observations.
Every eight new observations form a chunk, whose visual content
is summarized. Two frames are retained across chunk boundaries
to preserve contextual continuity. Every five pending chunk
summaries are compressed into a new long-term block and appended
to the existing memory. Consequently, recent experience remains
visually detailed, while earlier experience is represented through
compact descriptions of observed layouts, landmarks, and their
relationships.

Temporal summaries are generated without access to the navigation instruction, separating observational descriptions from task-conditioned judgments. A subsequent reasoning stage combines these summaries with the instruction and execution feedback. The temporal history is retained throughout subgoal transitions, while the completion assessment uses evidence associated with the active subgoal.

\subsubsection{Spatial Memory}
Spatial memory maintains a geometric map, the observed agent
trajectory, and a committed navigation target.
\begin{equation}
    m_t^{\mathrm{sp}} =
    \left(M_t,\tau_t,q_t\right).
\end{equation}
Here, $M_t$ is a ground-plane occupancy map, $\tau_t$ is the
trajectory accumulated from observed poses, and $q_t$ is the
active world-coordinate target, or $\varnothing$ when no
target is committed.

Depth observations are back-projected into world coordinates
using the camera calibration:
\begin{equation}
    \mathbf{x}_{t,u}^{w}
    =
    P_t
    \operatorname{BackProject}\!\left(u,D_t(u),K\right),
    \label{eq:backproject}
\end{equation}
where $\operatorname{BackProject}$ returns a homogeneous
camera-coordinate point for pixel $u$. The ground-plane
projections of these points update accumulated free-space
and obstacle evidence:
\begin{equation}
    M_t =
    \operatorname{Fuse}\!\left(M_{t-1},D_t,P_t,K\right).
\end{equation}
Each cell is classified as free, occupied, or unknown.
The resulting map supports obstacle-aware path planning
and frontier exploration. The two memories are coupled through pose anchoring. Each temporal
entry records the pose at which it was captured, and each summary
records the trajectory segment it covers. Temporal memory retains observed landmarks and their semantic relationships, complementing the geometric map.

\subsection{Memory-Conditioned Navigation Policy}
\label{sec:policy}

Under the POMDP formulation, the agent selects actions from its internal memory rather than the complete environment state. Section~\ref{sec:problem_formulation} defines this memory-conditioned policy, and Eq.~(\ref{eq:structured_memory}) specifies our representation. We implement the policy through high-level intent selection and geometric control, separating the choice of a navigation target from the primitive actions used to reach it:
\begin{equation}
    a_t = \kappa(M_t, P_t, q_t),
    \qquad
    q_t \sim \pi_\theta(\cdot \mid h_t).
    \label{eq:policy}
\end{equation}

The intent policy $\pi_\theta$ is implemented by a reasoning module
\begin{equation}
    (c_t, \eta_t) = \mathcal{J}_\theta\!\left(m_t^{\mathrm{task}}, m_t^{\mathrm{temp}}, m_t^{\mathrm{sp}}\right),
\end{equation}
whose two outputs address task progress and spatial intent. The completion flag $c_t \in \{0,1\}$ indicates whether the active subgoal $g_{k_t}$ has been achieved, and \texttt{STOP} is issued when $c_t = 1$ at $k_t = N$. The intent $\eta_t \in \mathcal{W}_t \cup \{\texttt{PREVIEW}\}$ either selects a target waypoint or requests a panoramic preview. 

The current egocentric observation and retained memory may not provide enough evidence to select a waypoint toward the active subgoal, for example when a referenced landmark lies outside the field of view. The reasoning module can therefore select \texttt{PREVIEW}, which rotates the agent in place to acquire additional views $\mathcal{V}_t=\{v_t^{(i)}\}_{i=1}^{n}$. These observations enter temporal memory and expand the waypoint candidate set to the surrounding area.

Previewing provides an opportunity to acquire missing evidence, but does not guarantee that the evidence resolves the current query or is correctly interpreted and retained. Under an idealized additive evidence update, additional facts can reduce the set of hypotheses consistent with the existing premises; Appendix~\ref{sec:semantic_information} formalizes this relation in Eq.~(\ref{eq:additive_evidence}). Our implemented memory update is not assumed to satisfy lossless evidence accumulation.

When a waypoint is predicted in image coordinates, Eq.~(\ref{eq:backproject}) is used to obtain its world-coordinate location for storage as the committed target in spatial memory. The target persists across steps until it is released by the memory update below, so a single intent spans multiple primitive steps. The controller $\kappa$ plans an obstacle-aware path to $q_t$ on the occupancy map $M_t$ and emits the next primitive action $a_t \in \mathcal{A}_{\mathrm{prim}}$.

After executing $a_{t-1}$, the agent receives $v_t$ and updates its memory as $h_t = f_{\mathrm{update}}(h_{t-1}, a_{t-1}, v_t)$. The update decomposes over the three memory components,
\begin{equation}
    f_{\mathrm{update}} = \left(f^{\mathrm{task}}, f^{\mathrm{temp}}, f^{\mathrm{sp}}\right),
\end{equation}
each of which consumes a different part of the input.

For task memory, the instruction $G$ and subgoals $\mathcal{G}$ are fixed after initialization, where $f_{\mathrm{init}}$ decomposes $G$ into $\mathcal{G}$ and sets $k_0 = 1$. Only the stage index changes, $k_t = \min(k_{t-1} + c_{t-1}, N)$, where $c_{t-1}$ is the completion flag computed by $\mathcal{J}_\theta$ from $h_{t-1}$. Hence $f^{\mathrm{task}}$ depends on $h_{t-1}$ alone.

For temporal memory, let $\delta_t$ denote the translation, rotation, and elevation change computed from $P_{t-1}$ and $P_t$. The recent window and execution history are updated as
\begin{equation}
    W_t = \operatorname{Last}_8\!\left(W_{t-1} \cup \{(I_t, \delta_t, P_t)\}\right),
    \qquad
    E_t = E_{t-1} \cup \{(a_{t-1}, \delta_t)\}.
\end{equation}
When eight new observations have accumulated since the last chunk, they form a chunk $B_j$, which also includes the last two frames of $B_{j-1}$ for continuity, and its summary is appended together with the trajectory segment it covers,
$C_t' = C_{t-1} \cup \{(\operatorname{Sum}(B_j), \tau_{[s_j, e_j]})\}$,
where $[s_j, e_j]$ is the step range of $B_j$; otherwise $C_t' = C_{t-1}$. Pending summaries are compressed into long-term memory once five have accumulated:
\begin{equation}
    (C_t, L_t) =
    \begin{cases}
        \left(\varnothing,\; L_{t-1} \cup \{\operatorname{Comp}(C_t')\}\right), & |C_t'| = 5,\\
        \left(C_t',\; L_{t-1}\right), & \text{otherwise}.
    \end{cases}
\end{equation}
The summaries are designed to retain observations relevant to later decisions while reducing the amount of history presented to the reasoning module.

For spatial memory, the map and trajectory integrate the new depth and pose, and the committed target is released once it is no longer valid:
\begin{equation}
    M_t = \operatorname{Fuse}(M_{t-1}, D_t, P_t, K),
    \qquad
    \tau_t = \tau_{t-1} \cup \{P_t\},
\end{equation}
\begin{equation}
    q_t =
    \begin{cases}
        \varnothing, & \operatorname{Done}(q_{t-1}),\\
        q_{t-1}, & \text{otherwise},
    \end{cases}
\end{equation}
where $\operatorname{Done}(\cdot)$ holds when the target is reached, passed, or invalidated by stagnation (detected from $E_t$) or by exceeding its execution budget.

Each step thus proceeds in three stages: $f_{\mathrm{update}}$ integrates the new observation into memory; if $q_t = \varnothing$, the intent policy commits a new target, and $\kappa$ emits the next primitive action toward it. Following the design rationale in Section~\ref{sec:design_rationale}, the agent uses memory-based reasoning to select targets and can request additional observations to support its decisions. This perception--reasoning--action loop repeats until \texttt{STOP} is issued.

\subsection{Action Abstraction}

We compare a \textbf{primitive action space} with an \textbf{intent action space}, both conditioned on the same observation $v_t$ and memory representation $h_t$. The two differ only in action abstraction: primitive actions specify low-level motion commands, whereas intent actions specify spatial targets that a geometric controller executes.

In the primitive action space, the policy selects $a_t \sim \pi_\theta(\cdot \mid h_t)$ from $\mathcal{A}_{\mathrm{prim}} = \{\texttt{FORWARD}, \texttt{LEFT}, \texttt{RIGHT}, \texttt{STOP}\}$, where \texttt{FORWARD} translates by a fixed distance and \texttt{LEFT}/\texttt{RIGHT} rotate by a fixed angle. Although this action space is small, reaching a spatial target requires a long sequence of such decisions, and the number of possible sequences grows as $|\mathcal{A}_{\mathrm{prim}}|^T$ over $T$ steps.

In the intent action space, the policy instead selects a waypoint $q_t \sim \pi_\theta(\cdot \mid h_t)$ from $\mathcal{A}_{\mathrm{intent}} = \mathcal{W}_t \cup \{\texttt{STOP}\}$, where $\mathcal{W}_t \subseteq \mathbb{R}^2$ is either a set of $K$ generated candidates or, for direct prediction, the local ground plane (image-space predictions are projected using depth and camera calibration). A geometric controller then produces the primitive actions $\mathbf{a}_t = \kappa(M_t, P_t, q_t)$, where $M_t$ is the accumulated map and $P_t$ the agent pose.

The key difference lies in the effective decision horizon. Per decision, $|\mathcal{A}_{\mathrm{intent}}| = K+1$ can exceed $|\mathcal{A}_{\mathrm{prim}}| = 4$. However, if a segment of $T$ primitive steps is covered by $N$ intents, with the $i$-th intent executing $\ell_i$ steps, then
\begin{equation}
    \frac{H_{\mathrm{intent}}}{H_{\mathrm{prim}}} = \frac{N}{T} = \frac{1}{\bar{\ell}} < 1,
    \qquad
    \bar{\ell} = \frac{1}{N}\sum_{i=1}^{N} \ell_i > 1.
\end{equation}
When an intent spans multiple primitive steps on average, intent prediction reduces the number of target-selection decisions required for a trajectory.

Consistent with this analysis, our method performs substantially better with waypoint prediction and selection than with primitive-action prediction. Table~\ref{tab:hsgm_vs_ours_token_comparison} reports a larger input-token load per decision for HSGM than for our method. This observation motivates reducing the context burden on smaller models, but does not isolate context length as the cause of the performance gap. Memory representation, model capability, action interfaces, and execution settings may also affect navigation performance.

\section{Evaluation Protocol}
We evaluate navigation on R2R-CE~\citep{krantz_vlnce_2020} using success rate (SR), success weighted by path length (SPL), navigation error (NE), oracle success rate (OSR), and normalized dynamic time warping (nDTW)~\citep{ilharco2019general}. We also report relative success rate, $\mathrm{RSR}=\mathrm{SR}/\mathrm{OSR}$, when OSR is positive and both metrics use the same cohort with complete oracle-success coverage. RSR measures successful stopping conditional on reaching the goal region; it does not separately identify perception, progress-estimation, or stopping errors.

LAW2 complements these metrics with paired instruction interventions. Each transformed instruction is compared with its original counterpart under the same scene, start pose, and execution settings:
\begin{itemize}
    \item \textsc{FLIP} reverses the initial turn command in 82 episodes to measure responsiveness to the supplied direction.
    \item \textsc{GOAL-ONLY} retains the final annotated sub-instruction in 174 episodes to measure dependence on route information. Success remains defined by the original target location.
    \item \textsc{PARAPHRASE} generates four intended meaning-preserving variants for each of 200 instructions: same register, terse, colloquial, and landmark re-description.
\end{itemize}
Appendix~\ref{app:law2} specifies cohort selection, transformations, validation, and scoring.

\begin{figure}[htbp]
    \centering
    \includegraphics[width=0.7\textwidth]{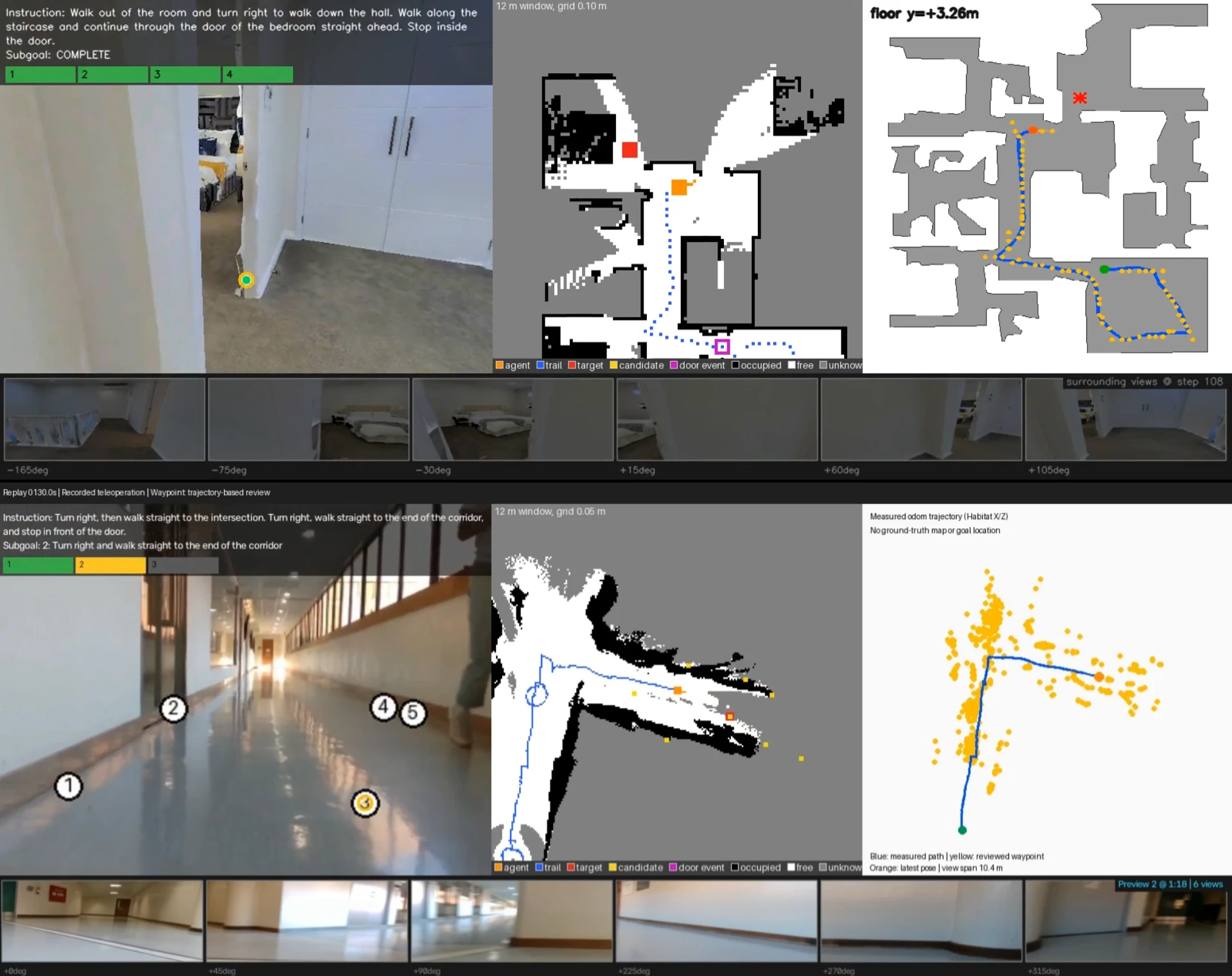}
    \caption{Qualitative results on R2R (top) and in real-world experiments (bottom).}
    \label{fig:QualitativeResults}
\end{figure}

\section{Experiments}
\subsection{Simulation Experiment}
Our method uses JoyAI-VL-Interaction-8B~\citep{yao2026joyai} as the observer, which summarizes visual observations into temporal memory, and Qwen3-VL-8B~\citep{bai2025qwen3} as the reasoner. We compare against both post-training and zero-shot baselines. The zero-shot baselines differ in how they use the LLM; for example, CA-Nav uses it only for subgoal decomposition. For the Qwen3-based comparisons, we retain each baseline's implementation and use the backbone configurations listed in Table~\ref{tab:r2r_val_unseen}. These comparisons share an 8B-scale language backbone but differ in auxiliary models, observation configurations, and action interfaces. All experiments run on a server with 8 NVIDIA A100 GPUs.

\subsection{Real-World Experiment}
We deploy our method and AwareVLN on a Unitree Go2 quadruped robot in corridor and office environments. Both models run on the same A100 server, which communicates with the Go2 remotely. The robot streams its onboard observations to the server and executes the returned actions. We reconstruct each environment with FAST-LIVO2~\citep{zheng2024fast} and localize the robot within the reconstructed map, which provides the pose $\hat{P}_t$ during navigation and the reference positions used for evaluation.

\subsection{Main Results}
Table~\ref{tab:r2r_val_unseen} shows that our complete system achieves SR of 0.263 and SPL of 0.196, the highest among the evaluated Qwen3-based zero-shot baselines. Post-trained models achieve higher SR and SPL in simulation. Our RSR of 0.802 indicates that a larger fraction of episodes reaching the goal region end successfully than for the other evaluated Qwen3-based zero-shot methods. This comparison concerns complete systems with different perception modules and action interfaces.

Adding temporal memory to CA-Nav increases SPL from 0.097 to 0.126 and nDTW from 0.162 to 0.225, while SR sligtly changes from 0.243 to 0.247. Average path length decreases from 27.82 to 23.50\,m. These results provide additional evidence that retained observation history can improve path efficiency, although the change in endpoint success is small.

In the two real-world environments, our system obtains higher SR and lower NE than AwareVLN(Table~\ref{tab:real-world}). This result demonstrate a better generalization ability that our method have. Figure~\ref{fig:QualitativeResults} is the qualitative results which illustrates simulation and real-world trajectories and task relevant elements visualization.

\subsection{Ablation Study}
Table~\ref{tab:ablation_results} reports the ablation results. We evaluate the proposed memory designs by replacing each component with a minimal baseline. For \textit{w/o Temporal}, the reasoner receives a sliding window of the eight most recent raw images, without historical observation reports and hierarchical summaries. For \textit{w/o Spatial}, the agent retains only its pose trajectory and the currently committed target. Waypoint candidates are generated from the current observatio, and the agent moves directly toward the selected target without an accumulated occupancy map. The \textit{w/o S\&T} variant combines both replacements. Task memory is retained in all variants, and preview remains enabled in the memory ablations. 

Removing temporal memory reduces SR from 0.263 to 0.190, SPL from 0.196 to 0.112, and nDTW from 0.379 to 0.201. Average path length increases from 16.9 to 39.1\,m and steps from 131 to 292, even though OSR rises from 0.328 to 0.475. The agent therefore reaches the goal region more frequently during longer trajectories but converts fewer visits into successful termination. This pattern is consistent with historical evidence supporting task progress and stopping decisions.

Removing the spatial component reduces SR to 0.205 and SPL to 0.142. Removing both spatial and temporal components yields SR of 0.185 and nDTW of 0.202. Task memory is retained in every variant, so these comparisons assess temporal and spatial components within the task-conditioned architecture.

Removing \texttt{PREVIEW} reduces SR from 0.263 to 0.190 while leaving nDTW nearly unchanged (0.379 versus 0.378). Token consumption also decreases. This tradeoff supports the utility of acquiring additional views within our system. Separately, the intent-action variant achieves higher SR and SPL than the primitive-action variant (0.263/0.196 versus 0.125/0.085), supporting the use of waypoint selection and geometric control to execute memory-conditioned decisions.

\subsection{Instruction-Intervention Analysis}
In \textsc{FLIP}, our initial-turn follow rate changes from 0.829 to 0.707, compared with 0.915 to 0.439 for AwareVLN and 0.646 to 0.268 for CA-Nav (Table~\ref{tab:FLIPExperiment}). Our system has the highest mean follow rate across the paired conditions. These results show differences in responsiveness to a changed direction command on the selected episodes.

In \textsc{GOAL-ONLY}, our SR decreases from 0.293 to 0.181, while nDTW increases from 0.354 to 0.390 and path length decreases from 19.37 to 11.25\,m (Table~\ref{tab:GoalOnlyExperiment}). AwareVLN shows decreases in both SR and nDTW, whereas CA-Nav shows lower SR with a small increase in nDTW. Thus, route-information removal affects endpoint success and original-route fidelity differently, and the zero-shot method tends to act like a human to explore with higher nDTW.

Under \textsc{PARAPHRASE}, our SR ranges from 0.240 to 0.280 around the original value of 0.270, while nDTW decreases from 0.387 to 0.350--0.365 across all four variants (Table~\ref{tab:paraphrase_arms}). Endpoint success is therefore relatively stable in the reported point estimates, whereas route fidelity changes. For AwareVLN, A4 reduces SR from 0.590 to 0.500 and nDTW from 0.621 to 0.566, indicating sensitivity to landmark re-description.

\begin{table*}[t]
\centering
\caption{Results on R2R-CE val-unseen. Bold: best within the post-training group, and best among
Qwen3(-VL)-8B-based methods in the zero-shot group.}
\label{tab:r2r_val_unseen}
\begingroup
\small
\setlength{\tabcolsep}{2.5pt}
\renewcommand{\arraystretch}{1.05}

\resizebox{\linewidth}{!}{%
\begin{tabular}{lllrrrrrrrr}
\toprule
Method & Navigation Backbone & Obs.
& SR $\uparrow$
& SPL $\uparrow$
& nDTW $\uparrow$
& OSR $\uparrow$
& NE $\downarrow$
& PL 
& Steps 
& RSR $\uparrow$ \\
\midrule
\multicolumn{11}{l}{\textit{Post-training}} \\
\midrule
Aware VLN\textsuperscript{c}~\cite{guo2026awarevln} & AwareVLN & Mono
& 0.647 & 0.563 & 0.656 & 0.744
& 4.10 & 12.79 & 107 & 0.869 \\

Aware VLN\textsuperscript{c}~\cite{guo2026awarevln} & AwareVLN (4-bit) & Mono
& 0.652 & 0.569 & \textbf{0.660} & \textbf{0.760}
& \textbf{3.89} & \textbf{12.69} & 110 & 0.858 \\

InternNav\textsuperscript{c}~\cite{wei2026ground} & InternVLA & Mono
& 0.585 & 0.538 & -- & 0.659
& 4.70 & -- & 99 & 0.888 \\

LightNav\textsuperscript{c}~\cite{wang2026lightnav} & LightNav & Mono
& \textbf{0.680} & \textbf{0.625} & 0.633 & 0.722
& 3.99 & -- & 84 & \textbf{0.942} \\

\midrule
\multicolumn{11}{l}{\textit{Zero-shot}} \\
\midrule
CA-Nav\textsuperscript{b,c}~\cite{chen2025constraint}
& Stack\textsuperscript{a} & Mono
& 0.243 & 0.097 & 0.162 & \textbf{0.479}
& 7.71 & 27.82 & 215 & 0.507 \\

CA-Nav + Temporal Memory\textsuperscript{b,c}~\cite{chen2025constraint}
& Stack\textsuperscript{a} & Mono
& 0.247 & 0.126 & 0.225 & 0.423
& 7.57 & 23.50 & 184 & 0.584 \\

GC-VLN\textsuperscript{b,c}~\cite{yin2025gc} & DeepSeek-R1 & Mono
& 0.141 & 0.066 & -- & 0.289
& 8.31 & 42.10 & 366 & 0.488 \\

\addlinespace[2pt]
SmartWay\textsuperscript{c}~\cite{shi2025smartway} & Q3VL-8B & Pano
& 0.178 & 0.134 & 0.380 & 0.361
& \textbf{7.39} & 15.81 & 12 & 0.493 \\

SmartWay~\cite{shi2025smartway} & GPT-4o & Pano
& 0.290 & 0.225 & -- & 0.510
& 7.01 & 13.09 & -- & 0.569 \\

\addlinespace[2pt]
Open-Nav\textsuperscript{c}~\cite{yuan2025opennav} & Qwen3-8B-FP8 & Pano
& 0.107 & 0.093 & \textbf{0.430} & 0.148
& 7.87 & 7.60 & 6 & 0.723 \\

Open-Nav~\cite{yuan2025opennav} & Llama-3.1-70B-Instruct & Pano
& 0.160 & 0.129 & 0.450 & 0.230
& 7.25 & 8.07 & -- & 0.696 \\

Open-Nav~\cite{yuan2025opennav} & GPT-4 & Pano
& 0.190 & 0.161 & 0.458 & 0.230
& 6.70 & 7.68 & -- & 0.826 \\

\addlinespace[2pt]
HSGM\textsuperscript{c}~\cite{li2026bridging} & Q3VL-8B & 3-view
& 0.049 & 0.048 & 0.272 & 0.120
& 8.79 & \textbf{6.27} & 34 & 0.408 \\

HSGM\textsuperscript{c}~\cite{li2026bridging} & GPT-5 & 3-view
& 0.463 & 0.324 & 0.463 & 0.569
& 5.47 & 15.36 & 43 & 0.814 \\

\midrule
Ours (Primitive Actions) & Q3VL-8B + JoyAI & Mono
& 0.125 & 0.085 & 0.320 & 0.225
& 8.05 & 15.76 & 213 & 0.556 \\

Ours (Intent Actions) & Q3VL-8B + JoyAI & Mono
& \textbf{0.263} & \textbf{0.196} & 0.379 & 0.328
& 7.56 & 16.98 & 131 & \textbf{0.802} \\
\bottomrule
\end{tabular}%
}
\par\smallskip
\begin{minipage}{\linewidth}
\scriptsize
Q3VL-8B: Qwen3-VL-8B-Instruct. --: unavailable.
Mono: single egocentric view; Pano: 12-view $360^{\circ}$ panorama; 3-view: front, left, right. Reproductions keep the original view configuration.
\textsuperscript{a}GroundingDINO + RepViT-SAM + BLIP2-ITM + BLIP-VQA + Q3VL-8B.
\textsuperscript{b}LLM used only for subgoal decomposition, not step-by-step navigation.
\textsuperscript{c}Reproduced by us; other rows are from the original papers, with RSR computed from reported SR and OSR.
\end{minipage}
\endgroup
\end{table*}

\section{Conclusion}
We studied what information a zero-shot navigator should retain when using 8B-scale models. Our design combines instruction progress, pose-anchored observation history, and accumulated geometry in complementary memories. The complete system achieves the highest SR and SPL among the evaluated Qwen3-based zero-shot baselines, and the ablations show the utility of temporal history and spatial mapping with planning. Temporal-memory integration with CA-Nav and deployment on a Unitree Go2 for real world experiment provide further evidence of practical utility. LAW2 supplements endpoint metrics by exposing changes in directional responsiveness and route fidelity under instruction interventions.

Future work will explore alternative memory mechanisms, including native memory and test-time training, alongside improved methods for extracting and compressing task-relevant information.

\subsection*{AI use statement}
Generative AI tools were used to construct and review the PARAPHRASE portion of LAW2. A language model generated four variants of each of 200 source instructions, and model-based review assessed semantic equivalence and flagged changes in directions and referents. Appendix~\ref{app:law2_paraphrase} describes generation, validation, and review. We also used generative AI tools to improve manuscript readability. The authors reviewed the AI-assisted content and take responsibility for the manuscript and reported results.

\subsection*{Reproducibility statement}
The paper specifies the memory representation, action interfaces, component ablations, and instruction-intervention protocols. Appendix~\ref{app:law2} provides cohort-selection rules and scoring details. We will release the implementation and accompanying configuration and benchmark records after acceptance.


\bibliography{iclr2027_conference}
\bibliographystyle{iclr2027_conference}
\appendix
\section{Additional Experiment Results}
There are some additional experiment results including ablation study, token assumption experiment, and LAW2 benchmark experiments.

\begin{table*}[htbp]
  \centering
  \caption{Ablation study evaluation of model variants across navigation performance, path efficiency, step budget, and average token consumption per episode.}
  \label{tab:ablation_results}
  \vspace{0.3em}
  \footnotesize 
  \renewcommand{\arraystretch}{0.85} 
  \setlength{\tabcolsep}{5pt} 
  \begin{tabular}{l ccccc}
    \toprule
    \textbf{Model Variant} & \textbf{Full} & \textbf{w/o Spatial} & \textbf{w/o Temporal} & \textbf{w/o S\&T} & \textbf{w/o Preview} \\
    \midrule
    \multicolumn{6}{l}{\textbf{\textit{Navigation Performance}}} \\
    SR \(\uparrow\)         & \textbf{0.263} & 0.205 & 0.190 & 0.185 & 0.190 \\
    SPL \(\uparrow\)        & \textbf{0.196} & 0.142 & 0.112 & 0.139 & 0.151 \\
    OSR \(\uparrow\)        & 0.328          & 0.345 & 0.475 & \textbf{0.492} & 0.295 \\
    NE \(\downarrow\)      & \textbf{7.56}  & 8.48  & 9.31  & 8.46  & 8.05  \\
    \midrule
    \multicolumn{6}{l}{\textbf{\textit{Path \& Trajectory Efficiency}}} \\
    Path \(\downarrow\)     & 16.9           & 18.6  & 39.1  & 31.7  & \textbf{16.7} \\
    Steps \(\downarrow\)    & 131            & 173   & 292   & 300   & \textbf{130}  \\
    nDTW \(\uparrow\)       & \textbf{0.379} & 0.316 & 0.201 & 0.202 & 0.378         \\
    SDTW \(\uparrow\)      & \textbf{0.168} & 0.131 & 0.095 & 0.117 & 0.141         \\
    Reach 500 Steps \(\downarrow\) & 162 & 171 & 657 & 693 & \textbf{108} \\
    \midrule
    \multicolumn{6}{l}{\textbf{\textit{Budget \& Speed}}} \\
    SR@100 Step \(\uparrow\) & \textbf{0.185} & 0.115 & 0.100 & 0.090 & 0.145 \\
    SR@150 Step \(\uparrow\) & \textbf{0.210} & 0.140 & 0.110 & 0.150 & 0.180 \\
    Med. Steps \(\downarrow\) & 68             & 85    & 96    & 120   & \textbf{67}   \\
    \midrule
    \multicolumn{6}{l}{\textbf{\textit{Token Consumption per Episode}}} \\
    8B In \(\downarrow\)    & 640k & 819k & 1084k & 1134k & \textbf{523k} \\
    8B Out \(\downarrow\)   & 10.9k & 13.9k & 27.1k & 27.9k & \textbf{9.4k} \\
    JoyAI In \(\downarrow\) & 224k & 282k & \textbf{114k} & 119k & 200k       \\
    JoyAI Out \(\downarrow\)& 5.4k & 8.2k & 10.1k & 10.7k & \textbf{1.6k}      \\
    \bottomrule
  \end{tabular}
\end{table*}

\begin{table}[H]
  \centering
  \caption{Comparison of VLM call frequency and token consumption between HSGM and Ours, reported per call/step and per episode.}
  \label{tab:hsgm_vs_ours_token_comparison}
  \resizebox{\columnwidth}{!}{%
  \begin{tabular}{l c c c c c c}
    \toprule
    & \textbf{VLM Calls} & \multicolumn{2}{c}{\textbf{Per Call / Step}} & \multicolumn{3}{c}{\textbf{Per Episode}} \\
    \cmidrule(lr){3-4} \cmidrule(lr){5-7}
    \textbf{Method} & \textbf{/ Episode} & \textbf{Input} & \textbf{Output} & \textbf{Input} & \textbf{Output} & \textbf{Total} \\
    \midrule
    HSGM & 11.7 & $\approx$23.0k & 285 & $\approx$270k & $\approx$3.3k & $\approx$273.3k \\
    \midrule
    Ours & 131 & $\approx$6.0k & $\approx$114 & 864k & 16.3k & 880.3k \\
    \quad \textit{-- Qwen3-VL-8B} & 131 & $\approx$4.9k & $\approx$83 & 640k & 10.9k & 651.9k \\
    \quad \textit{-- JoyAI}       & 131 & $\approx$1.7k & $\approx$41 & 224k & 5.4k  & 229.4k \\
    \bottomrule
  \end{tabular}%
  }
  \par\smallskip
\end{table}

\begin{table}[H]
  \centering
  \caption{Real-world evaluation in the corridor and the office.
  \emph{Simple Instruction}: follow a short instruction and stop.
  \emph{Complex Instruction}: a multi-step instruction that requires a turn or passing a doorway before reaching the target.
  NE (m) is the navigation error; SR is the success rate within $2.0$\,m.}
  \label{tab:real-world}
  \footnotesize
  \setlength{\tabcolsep}{2pt}
  \resizebox{0.75\columnwidth}{!}{%
  \begin{tabular}{lcccccccc}
    \toprule
    & \multicolumn{4}{c}{Corridor} & \multicolumn{4}{c}{Office} \\
    \cmidrule(lr){2-5} \cmidrule(lr){6-9}
    & \multicolumn{2}{c}{\shortstack{Simple\\Instruction}}
    & \multicolumn{2}{c}{\shortstack{Complex\\Instruction}}
    & \multicolumn{2}{c}{\shortstack{Simple\\Instruction}}
    & \multicolumn{2}{c}{\shortstack{Complex\\Instruction}} \\
    \cmidrule(lr){2-3} \cmidrule(lr){4-5} \cmidrule(lr){6-7} \cmidrule(lr){8-9}
    Method & NE$\downarrow$ & SR$\uparrow$ & NE$\downarrow$ & SR$\uparrow$
    & NE$\downarrow$ & SR$\uparrow$ & NE$\downarrow$ & SR$\uparrow$ \\
    \midrule
    AwareVLN & 2.85 & 0.3 & 3.20 & 0.1 & 3.10 & 0.2 & 3.46 & 0.0 \\
    \textbf{Ours} & \textbf{0.68} & \textbf{0.8} & \textbf{0.84} & \textbf{0.8}
    & \textbf{0.91} & \textbf{0.6} & \textbf{1.07} & \textbf{0.4} \\
    \bottomrule
  \end{tabular}}
\end{table}

\begin{table}[H]
  \centering
  \caption{Initial-turn obedience on 82 selected episodes of
  R2R Val Unseen. ORIG uses the original instructions, while FLIP
  reverses only the selected direction word. Both conditions use
  the same v1-3 start orientation. Follow, Opposite, and Straight
  are fractions of all evaluated episodes; Straight includes
  straight or insufficient departures. MeanFollow and WordEffect
  aggregate the paired conditions. SR measures success at the
  original goal and is auxiliary. Aggregate values are approximately
  computed from the rounded rates shown.}
  \label{tab:FLIPExperiment}
  \footnotesize
  \setlength{\tabcolsep}{5pt}
  \begin{tabular}{llcccccc}
    \toprule
    Method & Dataset
    & Follow$\uparrow$
    & Opposite$\downarrow$
    & Straight
    & SR$\uparrow$
    & MeanFollow$\uparrow$
    & WordEffect$\uparrow$ \\
    \midrule
    \multirow{2}{*}{Ours}
      & ORIG & 0.829 & 0.134 & 0.037 & 0.286
      & \multirow{2}{*}{\textbf{0.7680}}
      & \multirow{2}{*}{\textbf{0.2955}} \\
      & FLIP & 0.707 & 0.220 & 0.073 & 0.143
      & & \\
    \midrule
    \multirow{2}{*}{AwareVLN}
      & ORIG & 0.915 & 0.073 & 0.012 & 0.646
      & \multirow{2}{*}{0.6770}
      & \multirow{2}{*}{0.1830} \\
      & FLIP & 0.439 & 0.549 & 0.012 & 0.463
      & & \\
    \midrule
    \multirow{2}{*}{CA-Nav}
      & ORIG & 0.646 & 0.268 & 0.085 & 0.256
      & \multirow{2}{*}{0.4570}
      & \multirow{2}{*}{-0.0063} \\
      & FLIP & 0.268 & 0.671 & 0.061 & 0.171
      & & \\
    \bottomrule
  \end{tabular}
\end{table}

\begin{table}[H]
  \centering
  \caption{GOAL-ONLY experiment on 174 episodes of R2R Val Unseen, retaining only the final annotated sub-instruction. SR is evaluated against the unchanged target location, nDTW uses the original reference trajectory.}
  \label{tab:GoalOnlyExperiment}
  \footnotesize
  \setlength{\tabcolsep}{14pt}
  \begin{tabular}{lcccccc}
    \toprule
    \multirow{2}{*}{Method} & \multicolumn{3}{c}{ORIG} & \multicolumn{3}{c}{GOAL-ONLY} \\
    \cmidrule(lr){2-4} \cmidrule(lr){5-7}
    & SR$\uparrow$ & nDTW$\uparrow$ & PL (m) & SR$\uparrow$ & nDTW$\uparrow$ & PL (m) \\
    \midrule
    Ours     & 0.293 & 0.354 & 19.37 & 0.181 & 0.390 & 11.25 \\
    AwareVLN & 0.624 & 0.653 & 13.21 & 0.353 & 0.481 & 11.89 \\
    CA-Nav   & 0.241 & 0.162 & 28.34 & 0.184 & 0.168 & 25.75 \\
    \bottomrule
  \end{tabular}
\end{table}

\begin{table}[H]
  \centering
  \caption{PARAPHRASE experiment on 200 episodes of R2R Val Unseen.
    A1: same register; A2: terse; A3: colloquial; A4: landmark re-description.
    NE is in meters. CA-Nav uses Qwen for instruction parsing.}
  \label{tab:paraphrase_arms}
  \footnotesize
  \setlength{\tabcolsep}{5pt}
  \begin{tabular}{llcccccc}
    \toprule
    Method & Instruction & SR$\uparrow$ & OSR$\uparrow$ & RSR$\uparrow$ & SPL$\uparrow$ & NE$\downarrow$ & nDTW$\uparrow$ \\
    \midrule
    \multirow{5}{*}{Ours}
      & ORIG & 0.270 & 0.350 & 0.771 & 0.155 & 8.15 & 0.387 \\
      & A1   & 0.260 & 0.340 & 0.765 & 0.149 & 7.98 & 0.357 \\
      & A2   & 0.240 & 0.330 & 0.727 & 0.123 & 8.19 & 0.365 \\
      & A3   & 0.260 & 0.340 & 0.765 & 0.136 & 8.06 & 0.350 \\
      & A4   & 0.280 & 0.360 & 0.778 & 0.145 & 8.67 & 0.351 \\
    \midrule
    \multirow{5}{*}{AwareVLN}
      & ORIG & 0.590 & 0.760 & 0.776 & 0.499 & 4.11 & 0.621 \\
      & A1   & 0.630 & 0.720 & 0.875 & 0.509 & 4.60 & 0.611 \\
      & A2   & 0.610 & 0.720 & 0.847 & 0.517 & 4.79 & 0.641 \\
      & A3   & 0.620 & 0.720    & 0.861    & 0.511    & 4.72   & 0.626 \\
      & A4   & 0.500 & 0.640    & 0.781    & 0.483    & 4.96   & 0.566 \\
    \midrule
    \multirow{5}{*}{CA-Nav}
      & ORIG & 0.260 & 0.500 & 0.520 & 0.127 & 6.60 & 0.156 \\
      & A1   & 0.250 & 0.510    & 0.490    & 0.113    & 6.95   & 0.141 \\
      & A2   & 0.280 & 0.530    &0.528     & 0.123    & 6.83   & 0.131 \\
      & A3   & 0.210 & 0.520 & 0.404 & 0.088 & 7.36 & 0.130 \\
      & A4   & 0.210 & 0.490    & 0.429    & 0.074    & 7.58   & 0.086 \\
    \bottomrule
  \end{tabular}
\end{table}

\begin{figure}[htbp]
    \centering
    \includegraphics[width=0.7\textwidth]{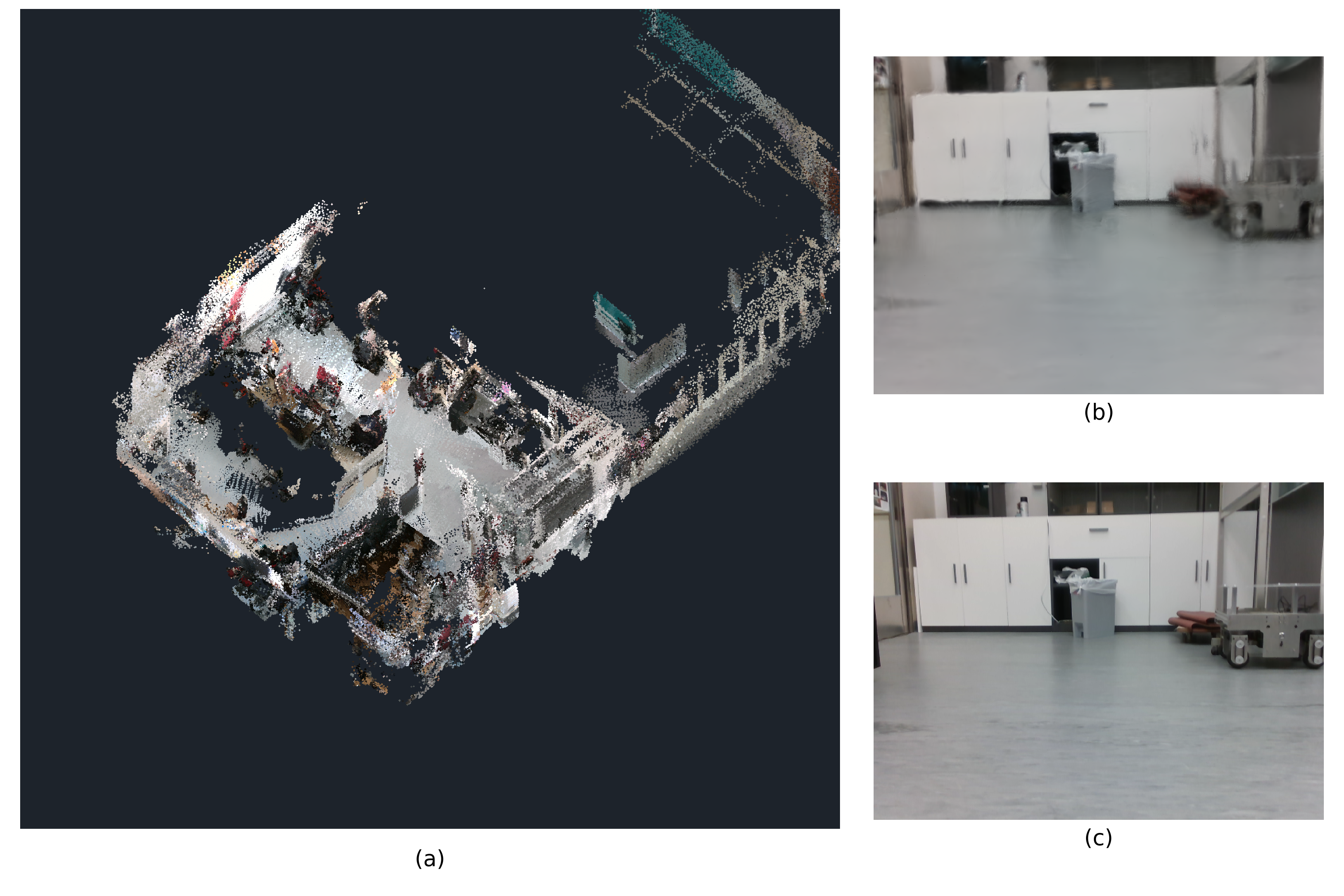}
    \caption{Qualitative results for the reconstructed office scene: (a) reconstructed point cloud; (b) image rendered using 3D Gaussian Splatting; and (c) original image.}
    \label{fig:ReconstructionOfOffice}
\end{figure}

\section{The LAW2 Benchmark}
\label{app:law2}

\subsection{Motivation and Evaluation Design}

LAW2, named after Asimov's second law~\citep{asimov2004robot}, evaluates instruction controllability through paired instruction interventions.

Each transformed instruction is compared with its matched \textsc{ORIG} instruction. Scene, start pose, goal, reference trajectory, model configuration, and execution settings are fixed within each pair.

\textsc{FLIP} tests directional responsiveness, \textsc{GOAL-ONLY} tests dependence on route information, and \textsc{PARAPHRASE} tests robustness to intended meaning-preserving rewrites. Expected effects differ: direction should change under \textsc{FLIP}, route fidelity may change under \textsc{GOAL-ONLY}, and behavior should remain stable under equivalent paraphrases.

\begin{table}[t]
\centering
\small
\begin{tabular}{lrl}
\hline
Transformation & Episodes & Instruction change \\
\hline
\textsc{FLIP} & 82 & Reverse the initial turn direction \\
\textsc{GOAL-ONLY} & 174 & Retain the final annotated segment \\
\textsc{PARAPHRASE} A1 & 200 & Same-register rewriting \\
\textsc{PARAPHRASE} A2 & 200 & Terse imperatives \\
\textsc{PARAPHRASE} A3 & 200 & Colloquial rewriting \\
\textsc{PARAPHRASE} A4 & 200 & Landmark re-description \\
\hline
\end{tabular}
\caption{\texttt{law2} cohorts. The four paraphrase styles share 200 source episodes and yield 800 rewrites; every condition has a matched \textsc{ORIG} control.}
\label{tab:law2_composition}
\end{table}

\subsection{Source Episodes and Instruction Alignment}

We use R2R-CE val-unseen~\citep{krantz_vlnce_2020} and FGR2R's human-annotated sub-instructions and viewpoint alignments~\citep{hong2020sub}.

\textsc{GOAL-ONLY} and \textsc{PARAPHRASE} draw from a fixed 200-episode manifest: 50 each for doorway, walking, stairs, and abstract stopping conditions (sampling seed 20260831). This is a stratified subset. \textsc{FLIP} candidates are selected separately from all 1,839 episodes.

Instructions are matched to FGR2R by trajectory identifier and normalized text. Annotation tokens are aligned to original character spans to preserve wording, punctuation, and capitalization during extraction.

Consecutive annotation segments ending at the same viewpoint are merged into navigation boundaries, yielding 589 segments across the 200 episodes. \textsc{PARAPHRASE} rewrites these merged segments; \textsc{GOAL-ONLY} extracts the final \emph{raw} segment.

\subsection{\textsc{FLIP}: Initial-Turn Obedience}
\label{app:law2_flip}

\subsubsection{Candidate Selection}
Testing the initial turn avoids conditioning evaluation on arrival at an intermediate intervention location.

An episode is eligible when:
\begin{enumerate}
\item Its first aligned instruction segment contains exactly one occurrence of \emph{left} or \emph{right}.
\item That occurrence expresses a movement command, such as ``turn left'' or ``take a right'', rather than a landmark relation such as ``the door on the left''.
\item The reference trajectory supports a turn in the commanded direction, with a magnitude between $45^\circ$ and $135^\circ$.
\item A probe in the opposite direction passes a local navigability check.
\end{enumerate}

Within the first 5\,m of horizontal path, we measure signed displacement angles relative to the initial heading at positions at least 0.5\,m from the start. The angle with greatest absolute magnitude must match the original direction word and lie within $45^\circ$--$135^\circ$.

We reflect this direction about the initial heading and place a probe 2\,m away. Its navigation-mesh snap distance must be at most 0.75\,m and vertical discrepancy below 0.5\,m. This checks local feasibility, not a complete alternative route to the goal.

The filter yields 94 candidates (53 left, 41 right). We shuffle each group with seed 0 and retain 41 per direction: 82 episodes across 11 scenes.

\subsubsection{Instruction Transformation Rules}
Only the selected direction word is reversed; all other text is retained.

Both conditions use the v1-3 start orientation. Exports from v1-2 records replace the start rotation accordingly. The paired control is \texttt{flip\_orig}; an ordinary \textsc{ORIG} run with a different orientation is invalid.

\subsubsection{Scoring Obedience}
Agent departures use the same geometry. Angles of at least
$30^\circ$ are scored as \emph{follow} or \emph{opposite} relative
to the supplied word; straight or insufficient departures remain
in the denominator as non-turning outcomes.

Let $F_i^r$ and $O_i^r$ indicate that episode $i$ follows or opposes
the supplied direction, respectively, in condition
$r\in\{\mathrm{orig},\mathrm{flip}\}$. Both indicators are zero
for non-turning outcomes. We compute
\begin{align}
\mathrm{MeanFollow}
&= \frac{1}{2N}\sum_{i=1}^{N}
\left(F_i^{\mathrm{orig}}+F_i^{\mathrm{flip}}\right),\\
\mathrm{WordEffect}
&= \frac{1}{4N}\sum_{i=1}^{N}
\left(
F_i^{\mathrm{orig}}+F_i^{\mathrm{flip}}
-O_i^{\mathrm{orig}}-O_i^{\mathrm{flip}}
\right).
\end{align}
WordEffect measures the net excess of following over opposing
to the supplied direction, scaled to $[-0.5,0.5]$. Taking the same
physical turning direction in both conditions contributes zero;
following both commands contributes $0.5$, and opposing both
contributes $-0.5$. Non-turning outcomes contribute zero.
Follow rates are also reported by condition and supplied direction.

Original-goal SR is auxiliary: the reversed command changes
the requested route, so goal success alone does not measure obedience.

\subsection{\textsc{GOAL-ONLY}: Dependence on Route Information}
\label{app:law2_goalonly}

\subsubsection{Cohort Selection}
From the 200-episode manifest, we exclude 24 episodes with one merged segment and two with exactly two merged segments ending in a bare stopping command, leaving 174 episodes. Selection is independent of agent performance.

\subsubsection{Strict Final-segment Extraction}
For the final raw FGR2R segment with character span $[a_i,b_i)$ in instruction $x_i$, we extract
\begin{equation}
x_i^{\mathrm{goal}} = \operatorname{strip}\bigl(x_i[a_i:b_i]\bigr).
\end{equation}
Only surrounding whitespace is removed, no text is restored, rewritten, or supplemented. Of the 174 retained segments.

\subsubsection{Evaluation}
We compare each transformed instruction with its original matched instruction on the same 174 episodes, keeping the target location and reference trajectory unchanged. SR measures successful stopping near the original target, and nDTW measures fidelity to the original route. Removing intermediate instructions may increase or decrease nDTW; these changes do not establish compliance with a uniquely specified alternative route.

\subsection{\textsc{PARAPHRASE}: Robustness to Expression}
\label{app:law2_paraphrase}

\subsubsection{Generation}
For each source instruction, the generator receives the original text, the ordered boundary-aligned segments. It returns one rewrite per segment for each of four styles:
\begin{itemize}
    \item A1, same register: reword the instruction while retaining its general style.
    \item A2, terse: use concise imperative expressions.
    \item A3, colloquial: use conversational phrasing, including discourse markers and hedging.
    \item A4, landmark re-description: use alternative expressions for the same intended physical referents while preserving distinguishing attributes and spatial relations.
\end{itemize}
Rewrites are required to preserve segment order, viewpoint alignment, and route boundaries. Each style is compared with its matched \textsc{ORIG} instruction. These are generation constraints, whose satisfaction is assessed by the validation and review procedures below.

\subsubsection{Automatic Validation}
Structural validation rejects outputs with missing styles, incorrect segment counts, empty segments, altered per-segment left/right sequences, or instruction lists that are entirely unchanged.

For A1--A3, a final segment that names a landmark must retain at least one matching content word, using a string-similarity threshold of 0.8. Segments consisting solely of a stopping command are exempt.

Additional checks flag newly introduced content words, A4 segments that retain all original content words, A2 rewrites longer than their source instructions, and A3 rewrites shorter than their source instructions. These flags trigger review rather than rejection: lexical overlap alone does not establish semantic equivalence.

All 800 rewrites passed the hard validation checks. Mean instruction lengths were 130.7 characters for \textsc{ORIG} and 138.6, 105.8, 186.2, and 159.4 characters for A1--A4, respectively.

\subsubsection{Review and Limitations}
Generation and semantic review used Claude Opus 5.5. Using the same model for both stages limits the independence of the assessment. 
The semantic review assessed all 800 rewrites together with 90 hidden control pairs, shuffled across six batches. Style labels and control identities were withheld. Each pair was classified as \emph{same}, \emph{weaker}, or \emph{different}, with separate flags for changes in direction and referent.

The controls comprised 30 identical pairs, 30 cross-episode pairs, and 30 direction-reversed pairs; all received the expected judgments. The initial review classified all A1--A3 rewrites and 199 of the 200 A4 rewrites as semantically equivalent. The remaining A4 rewrite replaced ``white flowers'' with ``pale blooms'' and was subsequently corrected.

\section{A Semantic Information Perspective on Embodied AI}
\label{app:semantic_information_theory_of_embodied_ai}

We describe how a receiver combines a communicated task message with its existing information, selects actions, and acquires further observations through environment interaction. The formulation separates information available for reasoning, realized computation, and satisfaction of a task specification.

\subsection{Available Information and Receiver Computation}
\label{sec:semantic_information}

\citet{lastras2026fundamental} study communication that enables a receiver to prove logical conclusions using its existing knowledge. Inspired by this receiver-centered perspective, we consider an initial message $G$ that communicates task requirements and may contain descriptive information. We model the receiver's computation explicitly and use a POMDP to describe its subsequent environment interaction. This formulation does not establish new communication bounds or assume that logical communication limits directly characterize task performance.

\subsubsection{Receiver Model}

We describe a receiver $\mathcal{R}$ through four elements: its internal configuration $\theta$, current internal information state $h_t$, information-processing mechanism $\mathcal{M}$, and output-selection settings $\lambda_t$. These specify relatively stable parameters, retained information available for the current computation, computational operations, and rules for selecting an output. This is a functional decomposition; it does not prescribe a particular architecture or memory organization.

Let $h_{\mathrm{init}}$ denote the receiver's initial information state and $o_t$ the observation received at step $t$. The initial message and subsequent interaction observations enter the receiver through
\begin{equation}
\begin{aligned}
h_0 &= U_{\theta,\mathcal{M}}^{\mathrm{init}}
(h_{\mathrm{init}},G,o_0), \\
h_t &= U_{\theta,\mathcal{M}}
(h_{t-1},a_{t-1},o_t), \qquad t\geq 1.
\end{aligned}
\label{eq:receiver_information_update}
\end{equation}
The message $G$ specifies what is requested and may provide descriptive claims. Observations provide information through interaction with the environment. The state $h_t$ is the receiver's internal state available for the current computation. The update $U_{\theta,\mathcal{M}}$ computes the current internal information state $h_t$ from the previous state $h_{t-1}$, the preceding action $a_{t-1}$, and the current observation $o_t$.

Computation on the internal information state and output selection are represented as
\begin{equation}
\begin{aligned}
z_t &= F_{\theta,\mathcal{M}}(h_t), \\
a_t &\sim D_{\lambda_t}(\cdot\mid z_t).
\end{aligned}
\label{eq:receiver_computation}
\end{equation}
Here, $F_{\theta,\mathcal{M}}$ performs computation on $h_t$ and produces the processed state $z_t$. We assume that $z_t$ contains the information required by $D_{\lambda_t}$. The functional notation describes deterministic processing, stochastic internal processing can instead be represented by conditional probability kernels.

\subsubsection{Semantic Sufficiency and Realized Computation}

Consider a fixed logical language and interpretation over a finite set $\mathcal{Z}$ of hypotheses about the environment and execution history. For a statement $F$, define
\begin{equation}
\operatorname{Ker}(F)
=
\{\zeta\in\mathcal{Z}:\zeta\models F\}.
\label{eq:semantic_kernel}
\end{equation}
The kernel contains the hypotheses compatible with $F$, and conjunction corresponds to intersection of kernels.

Let
\begin{equation}
K_t=\Phi_{\theta,\mathcal{M}}(h_t)
\label{eq:effective_premises}
\end{equation}
denote an idealized logical representation of the premises available to the receiver, including retained evidence and background knowledge invoked in reasoning. This is an analytical abstraction, not a guarantee that those premises are true. Descriptive claims from $G$ may contribute premises if retained and accepted as reliable; requested outcomes are not treated as facts that already hold.

For a task-relevant proposition $Q_t$, semantic sufficiency is expressed as
\begin{equation}
\varnothing\neq\operatorname{Ker}(K_t)
\subseteq\operatorname{Ker}(Q_t).
\label{eq:logical_sufficiency}
\end{equation}
Every hypothesis compatible with the premises then satisfies $Q_t$. The nonempty condition excludes inconsistent premises; correctness in the actual environment also requires the premises to hold there.

Semantic sufficiency does not guarantee that the receiver computes the conclusion or selects an appropriate action. Equation~(\ref{eq:receiver_computation}) describes realized computation, whose outputs need not be valid deductions. The same semantic content may also be easier or harder to use depending on its representation and the receiver's computational capabilities.

\subsubsection{Retained Information and New Evidence}

Under a fixed interpretation, making an existing semantic consequence explicit preserves the kernel:
\begin{equation}
\operatorname{Ker}(K_t\land Q_t)
=
\operatorname{Ker}(K_t),
\qquad K_t\models Q_t.
\label{eq:deductive_preservation}
\end{equation}
Additional observations may instead supply evidence that excludes previously compatible hypotheses. For additional interpreted evidence $E_t^{\mathrm{new}}$, an idealized additive update satisfies
\begin{equation}
\varnothing\neq
\operatorname{Ker}(K_t\land E_t^{\mathrm{new}})
\subseteq
\operatorname{Ker}(K_t),
\label{eq:additive_evidence}
\end{equation}
provided the combined premises remain consistent. The inclusion need not be strict, and additional evidence need not resolve the current query. Actual information-state updates may be lossy or revise earlier assumptions, so monotonic kernel reduction is not imposed on the receiver.

These relations distinguish retaining evidence, organizing it for computation, and acquiring further evidence. They do not prescribe how the receiver stores information or guarantee that it can identify which limitation is responsible for an unsuccessful decision.

\subsection{Interaction and Task Achievement}
\label{sec:intelligent_interaction}

For interaction with a partially observed environment, let $s_t\in\mathcal{S}$ be a Markov environment state, $a_t\in\mathcal{A}$ an executed action, and $o_t\in\mathcal{O}$ an observation. The receiver's internal state $h_t$ is represented separately. A POMDP specifies state transitions, observations, and a reward criterion~\cite{kaelbling1998planning}; we use a finite-horizon task-satisfaction objective below.

Given an initial-state distribution $\mu_0$ and initial-observation model $O_0$, initialization follows
\begin{equation}
s_0\sim\mu_0,
\qquad
o_0\sim O_0(\cdot\mid s_0).
\label{eq:interaction_initialization}
\end{equation}
Subsequent interaction follows
\begin{equation}
s_{t+1}\sim P(\cdot\mid s_t,a_t),
\qquad
o_{t+1}\sim O(\cdot\mid s_{t+1},a_t).
\label{eq:interaction_dynamics}
\end{equation}
Actions can change the physical environment, affect what is observed, or do both. Multi-step reasoning alone does not imply this interaction structure.

After receiving $o_{t+1}$, the receiver updates its information state according to Eq.~(\ref{eq:receiver_information_update}):
\begin{equation}
h_{t+1}
=
U_{\theta,\mathcal{M}}(h_t,a_t,o_{t+1}).
\label{eq:receiver_memory_update}
\end{equation}
For fixed initialization, mechanisms, and a specified rule for $\lambda_t$, these operations induce a policy $\pi_{\mathcal{R}}$. The receiver acts through its available information, without direct access to the complete environment state. Its internal state need not be an exact Bayesian belief state, and optimal inference or action selection is not assumed.

Let $Q_G$ specify the requirements used to evaluate execution for message $G$. For horizon $H$, let
\begin{equation}
\tau=(s_0,a_0,s_1,\ldots,a_{H-1},s_H),
\label{eq:task_trajectory}
\end{equation}
with early termination represented by an absorbing terminal state. Task-achievement probability is
\begin{equation}
J_{Q_G}(\mathcal{R})
=
\mathbb{E}_{\mathcal{R}}
\left[\mathbf{1}\{\tau\models Q_G\}\right]
=
\Pr_{\mathcal{R}}(\tau\models Q_G),
\label{eq:task_achievement}
\end{equation}
Here, $J_{Q_G}(\mathcal{R})$ is the receiver's task-achievement probability, $\tau\models Q_G$ means that the executed action set satisfies the task requirements, and $\mathbf{1}$ is one for satisfaction and zero otherwise. The expectation and probability are over trajectories induced by receiver computation and environment interaction, including randomness in initialization, transitions, observations, and action selection. The environment models, initialization distribution, horizon, and message are held fixed. This corresponds to zero intermediate rewards and a terminal task-satisfaction reward.

The local proposition $Q_t$ concerns a judgment during interaction, whereas $Q_G$ specifies what the executed action set must achieve. A receiver's conclusion or declaration of completion does not itself establish task satisfaction. The formulation therefore connects communicated information and receiver computation to outcomes produced through environment interaction.

\end{document}